\PassOptionsToPackage{dvipsnames}{xcolor}

\documentclass[sigconf]{acmart} 

\AtBeginDocument{%
  \providecommand\BibTeX{{%
    \normalfont B\kern-0.5em{\scshape i\kern-0.25em b}\kern-0.8em\TeX}}}

\usepackage[linesnumbered,ruled,vlined]{algorithm2e}

\usepackage[utf8]{inputenc}
\usepackage{tikz}
\usepackage{amsmath}
\usepackage{subcaption}
\usepackage{todonotes}
\usepackage{multirow}
\usepackage{enumitem}
\usepackage{pgfplots}
\usepackage{tabularx}

\usepgfplotslibrary{external,colormaps,groupplots,statistics}
\pgfplotsset{compat=1.8}
\usepackage{tikz}

\usetikzlibrary{shapes,arrows,fit,calc,positioning}
\usetikzlibrary{arrows.meta}

\definecolor{tabblue}{HTML}{1F77B4}
\definecolor{taborange}{HTML}{FF7F0E}

\tikzset{box/.style={draw, diamond, thick, text centered, minimum height=0.5cm, minimum width=0.5cm}}
\tikzset{leaf/.style={draw, rectangle, thick, text centered, minimum height=0.25cm, minimum width=0.5cm}}
\tikzset{line/.style={draw, thick, -latex'}}

\newcommand{\mytabrule}{\specialrule{.4pt}{0pt}{0pt}}
\newcommand{\NUMRUNS}{5}

\begin{document}

\title[Interpretable pipelines with evolutionarily optimized modules for RL tasks with visual inputs]{Interpretable pipelines with evolutionarily optimized modules for reinforcement learning tasks with visual inputs}


\author{Leonardo Lucio Custode}
\affiliation{%
  \institution{University of Trento}
  \city{Trento}
  \country{Italy}
}
\orcid{0000-0002-1652-1690}
\email{leonardo.custode@unitn.it}

\author{Giovanni Iacca}
\affiliation{%
  \institution{University of Trento}
  \city{Trento}
  \country{Italy}
}
\orcid{0000-0001-9723-1830}
\email{giovanni.iacca@unitn.it}

\renewcommand{\shortauthors}{Custode and Iacca}


\begin{abstract}
The importance of explainability in AI has become a pressing concern, for which several explainable AI (XAI) approaches have been recently proposed. However, most of the available XAI techniques are post-hoc methods, which however may be only partially reliable, as they do not reflect exactly the state of the original models. Thus, a more direct way for achieving XAI is through interpretable (also called glass-box) models. These models have been shown to obtain comparable (and, in some cases, better) performance with respect to black-boxes models in various tasks such as classification and reinforcement learning. However, they struggle when working with raw data, especially when the input dimensionality increases and the raw inputs alone do not give valuable insights on the decision-making process. Here, we propose to use end-to-end pipelines composed of multiple interpretable models co-optimized by means of evolutionary algorithms, that allows us to decompose the decision-making process into two parts: computing high-level features from raw data, and reasoning on the extracted high-level features. We test our approach in reinforcement learning environments from the Atari benchmark, where we obtain comparable results (with respect to black-box approaches) in settings without stochastic frame-skipping, while performance degrades in frame-skipping settings.
\end{abstract}



\maketitle


\section{Introduction}
\label{sec:introduction}

While the progress in AI continues to achieve new milestones, there is a growing concern on the need for understanding the decision-making process of AI models, especially in critical applications.
This awareness originated the subfield of \emph{explainable AI} (XAI), which has the goal to design tools to explain the decisions made by AI models, usually by means of post-hoc techniques.
However, while such techniques seem promising, they suffer from a fundamental issue: rather than reflecting the internal state of the explained model \cite{rudin_stop_2019, zhou_feature_2021}, they mainly focus on explaining its output, and how its behavior depends (either locally or globally) on the features of the problem at hand.
For this reason, another subfield of AI has been catching on in the past few years, namely that of \emph{interpretable AI} (IAI) \cite{barredo_arrieta_explainable_2020}.
Differently from XAI, IAI focuses on the development of \emph{inherently} interpretable models (also called ``explainable by design'' or ``glass-box'' models, as opposed to the traditional black-box ones), i.e., models that are directly understandable for humans without any post-hoc explanation.

Reinforcement learning (RL) is a particularly interesting setup for evaluating such models, since several real-world problems can be (and, in fact, have been) modelled as RL tasks \cite{dulac2019challenges}, for instance in robotics \cite{kormushev2013reinforcement}, autonomous driving \cite{lange2012autonomous}, unmanned aerial vehicles \cite{abbeel2007application}, production scheduling \cite{wang2005application}, resource allocation in the cloud \cite{barrett2013applying}, and medicine \cite{yu2021reinforcement,gottesman2019guidelines}. In all these fields, explainability is an issue not only from a technical standpoint, but also from a legal and, to some extent, ethical perspective.
The recent literature has proposed some seminal approaches for performing RL with interpretable models \cite{silva_optimization_2020,dhebar_interpretable-ai_2020}, e.g., based on evolutionary computation \cite{custode_co-evolutionary_nodate,custode_evolutionary_2021}. So far, these interpretable reinforcement learning (IRL) approaches have been mostly tested on relatively simple control RL tasks, such as those of the OpenAI gym benchmark \cite{brockman16openai}, on which they have obtained fairly good results. However, these methods are not expected to work well in RL tasks with raw input data, as in these contexts each variable alone (e.g., a pixel) may not be meaningful enough to take decisions.

In this work, we introduce the concept of \emph{interpretable pipelines} for tackling RL tasks with visual inputs.
An interpretable pipeline is a multi-agent system where each agent is an interpretable model with well-defined responsibilities, which communicates with the other agents in the pipeline.
We optimize such pipelines by means of a co-evolutionary approach, in which different evolutionary algorithms (EAs) run in parallel, each of which optimizes a single agent.
We test our approach on three different Atari games, where we observe that the proposed method is able to achieve satisfactory performance in deterministic settings (i.e., without frame-skipping). On the other hand, our approach is not able to achieve satisfactory performance in environments with stochastic frame-skipping (yielding higher uncertainty about the future), which provides some hints for future work.


The paper is structured as follows.
The next section makes a brief overview of the related work.
Section \ref{sec:method} explains the methodology used in our experiments.
In Section \ref{sec:results} we present the experimental setup and the results and, finally, in Section \ref{sec:conclusions} we draw the conclusions of this work.


\section{Related work}
\label{sec:related_work}

IRL has recently gained attention in the research community.
Silva et al., in \cite{silva_optimization_2020}, employed differentiable decision trees trained by means of the PPO algorithm \cite{schulman_proximal_2017}.
A differentiable decision tree is a decision tree that, instead of using binary conditions (also called \emph{hard} splits), uses \emph{soft} splits.
Each soft split is defined as $\sigma(x - v)$, where $\sigma$ denotes the sigmoid function.
Then, when computing the output for a given sample, the leaves are weighted according to the weights encountered during the path leading to that leaf, i.e., $\sigma(x - v_i)$ for the ``True'' branch and $(1 - \sigma(x - v))$ for the ``False'' branch.
This approach gives satisfactory results when using the differentiable version of the decision trees (which have low interpretability).
However, when the produced trees are discretized into decision trees with hard splits (which have high interpretability), significant losses in performance occur.

Another interesting approach to IRL has been proposed in \cite{dhebar_interpretable-ai_2020}.
Here, the authors introduced a methodology that optimizes decision trees with non-linear splits by using an EA.
The results show that the proposed approach works well in discrete-action settings.
However, the highly non-linear splits limit the interpretability of the solutions produced.

In \cite{custode_evolutionary_2021}, the authors proposed a methodology based on Grammatical Evolution (GE) \cite{goos_grammatical_1998} and Q-learning \cite{watkins_learning_1989} to produce decision trees that perform online learning. More specifically, the GE algorithm is used to optimize the inner splits of the decision trees, while Q-learning is used to learn the discrete actions for the leaves. This method was tested on three control tasks from the OpenAI gym benchmark, achieving state-of-the-art trade-offs in terms of performance and interpretability.
In \cite{custode_co-evolutionary_nodate}, this approach was further extended in order to handle RL tasks with continuous action spaces.
Here, the authors employed a co-evolutionary system based on two independent evolutionary processes: the first one, based on GE, optimizes the decision trees; the second one, based on an Estimation of Distribution Algorithm \cite{hauschild2011introduction}, optimizes pools of continuous actions.

As mentioned earlier, the main limitation of these approaches is that while they can be effective in tasks with a small number of high-level features, they are not expected to work in environments with high-dimensional, low-level features, such as images.
In fact, in the latter scenario, each input to the system does not provide significant information for the decision-making process. 
Moreover, even if one of those methods was able to obtain satisfactory performance by using a subset of the raw input data (i.e., a part of an image), making it generalize to other settings would be extremely hard.
Furthermore, applying those methods straightforwardly on the raw data would achieve limited interpretability.

Concerning this latter aspect, it is important to note that interpretability intended as a binary property is ill-defined.
For this reason, some works proposed approaches to quantitatively measure interpretability.

In \cite{virgolin_learning_2020}, the authors learned a metric of interpretability by training a regression model on the results of a survey. The resulting metric was:
\[
\mathcal{M}(\ell, n_o, n_{nao}, n_{naoc}) = 79.1 - 0.2\ell - 0.5n_o -3.4n_{nao} - 4.5n_{naoc}
\]
where:
\begin{itemize}[leftmargin=*]
    \item $\ell$ is the number of symbols in the formula;
    \item $n_o$ is the number of operations in the formula;
    \item $n_{nao}$ is the number of non-arithmetical operations;
    \item $n_{naoc}$ is the maximum number of consecutive compositions of non-arithmetical operations.
\end{itemize}
This metric is intended to lie in $[0, 100]$, where $0$ means ``non-interpretable'' and $100$ means ``interpretable''.
However, when applying this formula on large models, there is the possibility that the $\mathcal{M}$-score exceeds the bounds.
For this reason, in \cite{custode_evolutionary_2021,custode_co-evolutionary_nodate}, the authors rewrote this metric as:
\begin{equation}
\mathcal{M'}(\ell, n_o, n_{nao}, n_{naoc}) = -0.2 + 0.2\ell + 0.5n_o + 3.4n_{nao} + 4.5n_{naoc}
\label{eq:mprime}
\end{equation}
This version of the $\mathcal{M}$ metric works essentially as a complexity metric, and is defined in $[0, \infty)$.
In this case, a constant model has $\mathcal{M'} = 0$, which is the best possible value.
As the value of the $\mathcal{M'}$ moves away from $0$, the interpretability of the system decreases.

The idea of using complexity as a proxy for interpretability was also proposed in \cite{barcelo_model_2020}, where the authors stated that the computational complexity of a model can be used as a metric of interpretability as it directly resembles the number of operations that must be interpreted by humans.


\section{Method}
\label{sec:method}

To evolve interpretable pipelines for image-based RL tasks, we build on some of the aforementioned previous works from the literature \cite{custode_evolutionary_2021, custode_co-evolutionary_nodate,tang_neuroevolution_2020}.
In detail, our proposed system is an interpretable pipelines composed of two parts:
\begin{itemize}[leftmargin=*]
    \item a vision module, that is meant to process the input to extract a pre-defined number of features;
    \item a decision module, whose purpose is to decide which action to take, based on the features extracted by the vision module.
\end{itemize}

It is important to note that in this work, features represent high-level visual information (position of relevant objects), while decisions are actions taken by a decision tree (playing one of the Atari games considered in our experimentation). However, the proposed pipelines can be extended to tackle classification problems.

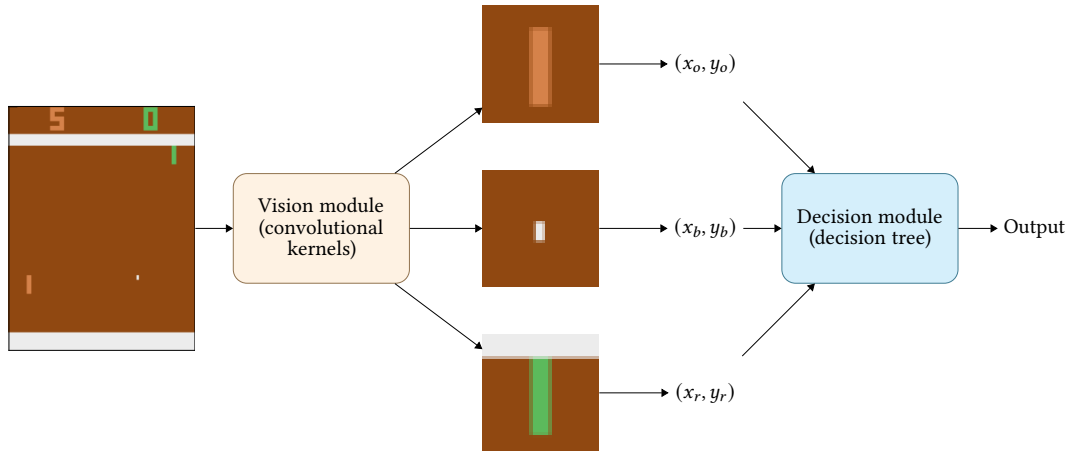
\begin{figure*}[ht!]
 \centering
 \resizebox{0.8\textwidth}{!}
 {
 \begin{tikzpicture}
 \node[inner sep=0pt] (frame) at (0,0)
    {\includegraphics[width=.4\columnwidth]{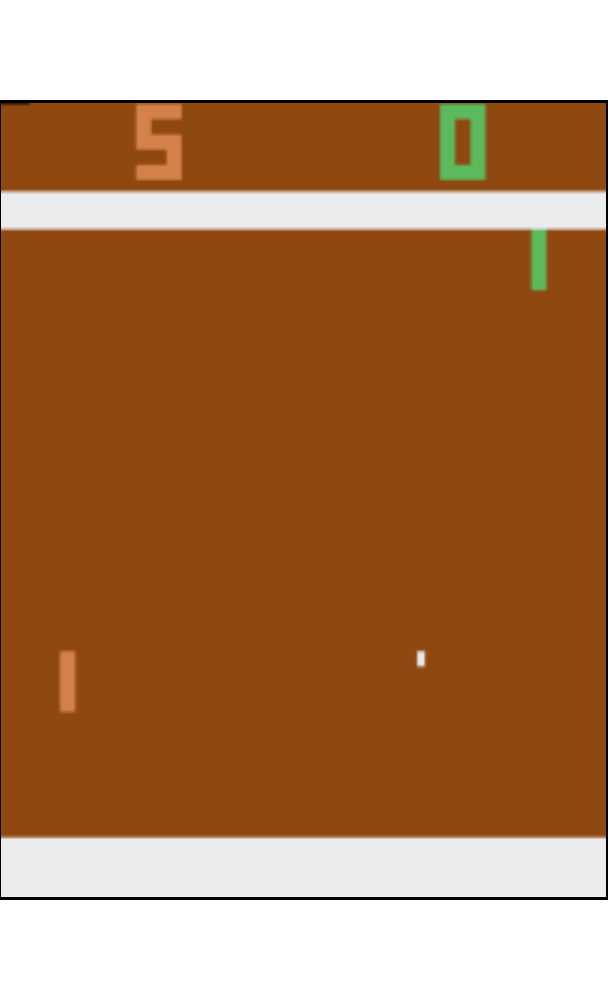}};
 
 \node[text width=3cm, draw=black, minimum width=3cm, minimum height=2cm, align=center, rounded corners=0.25cm, fill=BurntOrange!10, draw=BurntOrange!50!black] (vision) at (4, 0, 0) {\Large{Vision module\\(convolutional kernels)}};

 \node[inner sep=0pt] (opp) at (8,3)
    {\includegraphics[width=.25\columnwidth]{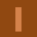}};
 \node[inner sep=0pt] (ball) at (8,0)
    {\includegraphics[width=.25\columnwidth]{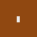}};
 \node[inner sep=0pt] (rack) at (8,-3)
    {\includegraphics[width=.25\columnwidth]{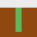}};

 \node[ draw=white,  minimum height=1.5cm, align=center, rounded corners=0.25cm, ] (opp_xy) at (11, 3, 0) {\Large$(x_{o}, y_{o})$};
 \node[ draw=white,  minimum height=1.5cm, align=center, rounded corners=0.25cm, ] (ball_xy) at (11, 0, 0) {\Large$(x_{b}, y_{b})$};
 \node[ draw=white,  minimum height=1.5cm, align=center, rounded corners=0.25cm, ] (rack_xy) at (11, -3, 0) {\Large$(x_{r}, y_{r})$};
 
 \node[text width=3cm, draw=black, minimum width=3cm, minimum height=2cm, align=center, rounded corners=0.25cm, fill=ProcessBlue!15, draw=ProcessBlue!50!black] (decision) at (14, 0, 0) {\Large{Decision module\\(decision tree)}};
 
 \node (output) at (17, 0, 0) {\Large{Output}};
 
 \draw[-Triangle] (frame.east) -- (vision);
 \draw[-Triangle] (vision) -- (opp);
 \draw[-Triangle] (vision) -- (ball);
 \draw[-Triangle] (vision) -- (rack);
 \draw[-Triangle] (rack) -- (rack_xy);
 \draw[-Triangle] (ball) -- (ball_xy);
 \draw[-Triangle] (opp) -- (opp_xy);
 \draw[-Triangle] (rack_xy) -- (decision);
 \draw[-Triangle] (ball_xy) -- (decision);
 \draw[-Triangle] (opp_xy) -- (decision);
 \draw[-Triangle] (decision) -- (output);
 \end{tikzpicture}
 }
 \caption{An example of the proposed pipelines (note that the frames are taken from the Pong environment).}\label{fig:pipeline}
\end{figure*}

A graphical representation of this kind of pipelines is shown in Figure \ref{fig:pipeline}. 
In the following subsections, we will first explain the details of the two kinds of modules, and then we will describe our co-evolutionary approach.

\subsection{Vision module}
\label{subsec:vision_module}

In \cite{tang_neuroevolution_2020}, the authors used a simplified self-attention module that ranks the patches of the image by importance.
Then, the coordinates of the $k$ most important patches are given in input to an LSTM network \cite{hochreiter1997long} that computes the decision to take, where $k$ is a predetermined parameter.
Similarly, in this work we use a vision module, whose purpose is to find the $k$ most important patches in the image, returning their coordinates.

However, in order to have better interpretability, rather than a self-attention module, in our vision module we employ $k$ convolutional kernels, each of which is supposed to detect a single entity of interest in the image.

Moreover, using $k$ distinct kernels instead of a single self-attention module allows us to have a fixed-order constraint on the input: e.g., given two kernels $k_A$ and $k_B$ which respectively handle the detection of two distinct objects $A$ and $B$, we are guaranteed that the decision module will receive in input the position of the two entities always in the same order.
In contrast, when using a self-attention module as done in \cite{tang_neuroevolution_2020}, we do not have such guarantees, and thus the decision module has to handle such cases of inputs in unpredictable order, thus requiring a more complex decision logic.

The pseudo-code that describes how the vision module works is shown in Algorithm \ref{alg:vision}. In the algorithm $R:\mathbb{R}^{h'\times w'}$ is the output of the convolution between the image and the kernel, which results in a matrix of size $h'\times w'$, where $h' = h_{img} - 2\lfloor{\frac{h_k}{2}}\rfloor$ and $w' = w_{img} - 2\lfloor{\frac{w_k}{2}}\rfloor$, and
$h_{img}$ is the height of the image, $h_k$ is the height of the kernel, $w_{img}$ is the width of the image, $w_k$ is the width of the kernel.

\begin{algorithm}
    \SetCommentSty{textnormal}
	\caption{Computation of the vision module}
	\label{alg:vision}
	\SetAlgoLined
	\KwData{an image, $\textbf{X}$}
	\KwData{a list of kernels, $\textbf{k}$}
	\KwResult{a list of coordinates, $\textbf{c}$}
	$\textbf{c} \gets []$\;
	\For{$k \in \textbf{k}$}{
		$R \gets convolution(\textbf{X}, k)$\tcp*[r]{apply kernel $k$ to \textbf{X}}
		$(x^*, y^*) \gets \underset{(x, y)}{argmax}(R[y, x])$\tcp*[r]{$(x, y)$ size of $\textbf{X}$}
		$\textbf{c} \gets \textbf{c} + [(x^*, y^*)]$\tcp*[r]{concatenate $\textbf{c}$}
	}
	\Return $\textbf{c}$\;
\end{algorithm}

\subsection{Decision module}
\label{subsec:decision_module}

The goal of the decision module is to perform ``reasoning'' on the coordinates of the most important patches of the input images and to take a decision on top of them.

To keep the interpretability of the pipelines high, we use an automatically synthesized decision tree as decision module.
This decision tree takes as input the list of coordinates computed by the vision module, thus it does not use the raw data of the whole image.

This module then returns a discrete action that can be directly sent to the environment, see the examples shown in Figure \ref{fig:bests} and the corresponding analysis reported in Section \ref{subsec:analysis}.

\subsection{Co-evolutionary process}
\label{subsec:coevolutionf}

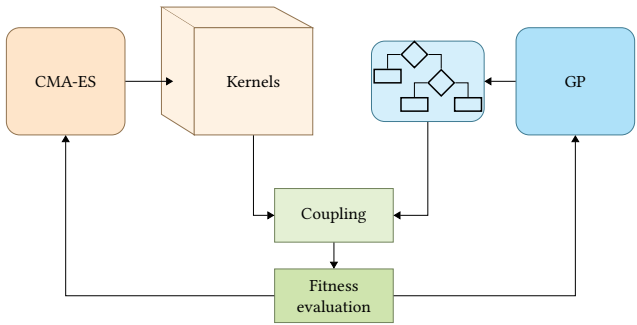
\begin{figure}[ht!]
 \centering
 \resizebox{\columnwidth}{!}
 {
 \begin{tikzpicture}
 \node[text width=2cm, fill=BurntOrange!20, draw=BurntOrange!50!black, minimum width=2cm, minimum height=2cm, align=center, rounded corners=0.25cm] (cmaes) at (-3.5,0,0) {CMA-ES};
 
 \draw[fill=BurntOrange!10, draw=BurntOrange!50!black] (0, 1, -1) -- (-2.1, 1, -1) -- (-1.1, 1, 0) -- (1.1, 1, 0) -- (0, 1, -1);
 \draw[fill=BurntOrange!10, draw=BurntOrange!50!black] (-2.1, 1, -1) -- (-2.1, -1, -1) -- (-1.1, -1, 0) -- (-1.1, 1, 0);
 \node[text width=2cm, fill=BurntOrange!10, draw=BurntOrange!50!black, minimum width=2cm, minimum height=2cm, align=center] (kernels) at (0,0,0) {Kernels};
 
 \node [minimum width=2.1cm, minimum height=1.5cm, fill=ProcessBlue!15, draw=ProcessBlue!50!black, rounded corners=0.25cm] (tree) at (3.25, 0, 0) {\ };
 \node [box] (root) at (3, 0.5, 0) {\ };
 \node [leaf, below=-0cm of root, xshift=-0.5cm] (l) {\ };
 \node [box, below=-0cm of root, xshift=+0.5cm] (r) {\ };
 \node [leaf, below=-0cm of r, xshift=-0.5cm] (rl) {\ };
 \node [leaf, below=-0cm of r, xshift=+0.5cm] (rr) {\ };
 
 \draw (root) -| (l) node [midway, above] () {\ };
 \draw (root) -| (r) node [midway, above] () {\ };
 \draw (r) -| (rl) node [midway, above] () {\ };
 \draw (r) -| (rr) node [midway, above] () {\ };

 \node[text width=2cm, fill=ProcessBlue!30, draw=ProcessBlue!50!black, minimum width=2cm, minimum height=2cm, align=center, rounded corners=0.25cm] (gp) at (6,0,0) {GP};
 
 \node[text width=2cm, fill=LimeGreen!20, draw=LimeGreen!50!black, minimum width=2cm, minimum height=1cm, align=center] (coupling) at (1.5,-2.5,0) {Coupling};
 
 \node[text width=2cm, fill=LimeGreen!40, draw=LimeGreen!50!black, minimum width=2cm, minimum height=1cm, align=center] (fitness) at (1.5,-4,0) {Fitness evaluation};
 
 \draw[-Triangle] (cmaes) -- (-1.5, 0, 0);
 \draw[-Triangle] (kernels.south) |- (coupling.west);
 \draw[-Triangle] (coupling.south) -| (fitness);
 \draw[-Triangle] (fitness) -| (cmaes);
 \draw[-Triangle] (gp) -- (tree);
 \draw[-Triangle] (tree) |- (coupling);
 \draw[-Triangle] (fitness) -| (gp);
 \end{tikzpicture}
 }
 \caption{Scheme of the proposed co-evolutionary process.}\label{fig:coevolution}
\end{figure}

To optimize the vision and decision modules adopted within the proposed pipelines, we employ a co-evolutionary approach \cite{popovici2012coevolutionary}.
In particular, we combine Covariance Matrix Adaptation Evolution Strategies (CMA-ES) \cite{hansen1996adapting} with Genetic Programming (GP) \cite{koza2005genetic}, as shown in Figure \ref{fig:coevolution}.
We use CMA-ES to evolve the parameters of the vision module, i.e., the weights of each kernel module. CMA-ES has been chosen for being one of the most robust algorithms for derivative-free optimization.
On the other hand, by using Genetic Programming (more specifically, strongly typed Genetic Programming \cite{montana1995strongly}), we evolve decision trees, as described below.

\subsubsection{Genetic Programming for evolving decision trees}
To evolve decision trees, we use two types of nodes: condition nodes and leaf nodes.
A condition is represented as a node with three child nodes: a comparison node and two nodes (either leaves or conditions).

A comparison node is composed of a node representing an operator (e.g., ``less than'', ``equal to'', and ``greater than'') that has two child nodes, encoding two expressions.

An expression node can be either a constant, a variable, or an arithmetical operation between expression nodes.

Since the interpretability of a decision tree crucially depends on the complexity of the conditions (i.e., more complex hyperplanes are hard to interpret, e.g., the ones presented in \cite{dhebar_interpretable-ai_2020}), not only we use a constant to limit the size of the tree, but we also employ a different constant to limit the depth of the conditions.

By doing so, we can control better the interpretability of the tree by allowing, for instance, deeper trees with simple conditions.

\subsubsection{Fitness evaluation}
In order to evaluate the quality of the individuals from both populations, we pair each individual with all the individuals of the other population.

We evaluate the pair on $e$ episodes, and we compute the average score (across episodes) for the pair $\bar{s}_{i,j}$, where $i$ is the index of the individual from the population of vision modules and $j$ is the index of the individual from the population of decision modules.

Then, we define the fitness of an individual as the maximum $\bar{s}$ that that individual obtained across all the pairings, i.e. $f_i = \underset{j}{max}(\bar{s}_{i, j})$ for vision modules and $f_j = \underset{i}{max}(\bar{s}_{i, j})$ for decision modules.

While using a different operator such as the mean (across pairings) seems more meaningful, from preliminary experiments we observed that using the mean leads to a stagnation of the co-evolutionary process.
We hypothesize that this is due to the fact that, by using the mean, the fitnesses are affected by the randomness in the evaluation phase, so that an individual that obtains medium-low scores in all the pairings may have a higher fitness than an individual that works very well combined with a specific individual of the other population but works poorly with all the other individuals.

\subsubsection{Reducing the number of evaluations}
\label{subsubsec:reducing_the_number_of_evaluations}

The computational cost (in terms of number of evaluations, where a single evaluation relates to an instance of the proposed pipelines) for the co-evolutionary process is $\mathcal{O}(p_c \cdot p_g \cdot g \cdot e)$, where $p_c$ is the population size for CMA-ES, $p_g$ is the population size for the Genetic Programming (assuming that each individual of one population is paired with all the individuals of the other population), $g$ is the number of generations, and $e$ is the number of episodes.

To reduce such cost, we propose a mechanism that evaluates the \emph{behavior} (i.e., the output) of each individual in the two populations, and avoids the evaluation of individuals whose behavior is too similar. At each generation, for both the vision and decision modules in the current populations of the co-evolutionary process, we give them in input a set of samples and we store their outputs. Then, for both populations (separately), we cluster these outputs by means of the DBSCAN algorithm \cite{schubert2017dbscan}. More specifically, the samples used for clustering are calculated as follows.
\begin{itemize}[leftmargin=*]
    \item \textbf{Vision modules} -- For these modules, we evaluate their behavior by giving them in input a set of $n_{vm}$ images sampled randomly from an episode used at the beginning of the evolutionary process. Then, we use the concatenation of all the outputs of each module as input samples for the clustering process.
    \item \textbf{Decision modules} -- For these modules, we evaluate their behavior by giving them in input $n_{dm}$ randomly sampled coordinates from the vision modules' output space. Then, we perform a one-hot encoding of the decision modules' outputs and, by concatenating all the one-hot encoded outputs for each module, we obtain the input samples for the clustering process. The reason underlying the one-hot encoding is related to the ``meaning'' of the input samples for the clustering process. In fact, if we used the raw outputs of the decision modules as input samples for clustering, we would give an implicit ``proximity'' meaning such that an action $a_i \in \mathcal{A}=\{a_0, \dots, a_i, a_{i+1}, a_{i+2}, \dots, a_A\}$ would be considered ``closer'' to action $a_{i+1}$ than $a_{i+2}$, while the meaning of such actions may not have such a ``similarity'' principle. Instead, by performing a one-hot encoding, the distance between two different actions (performed by different decision modules given the same input) is always constant, thus removing the bias associated to the raw outputs.
\end{itemize}
Once clustering has been performed, we evaluate only the centroids of each cluster (and the individuals not assigned to any cluster) and we assign, for each individual of the cluster, the same fitness.

The reason underlying the choice of the DBSCAN algorithm is due to the fact that this algorithm is based on the concept of \emph{density}, so that we can intuitively set the thresholds for considering two points as ``close''.
Moreover, this algorithm does not require to specify a pre-defined number of clusters, so this means that if we find a cluster, its points are close enough to be considered similar.
Finally, we do not use constant parameters for the distance threshold $\varepsilon$.
Instead, we use an initial $\varepsilon_0$, which, at each step, is multiplied by a scaling constant $\gamma \in [0, 1)$.
This scaling mechanism gives us the following properties.
At the beginning of the evolutionary process, since the diversity is high, we perform a coarse-grained clustering of the individuals so that we can significantly speed up the initial generations.
On later generations, the thresholds become increasingly smaller, so that even not-so-diverse individuals are evaluated separately: this, in turn, leads to a greater number of simulations, which allows us to discriminate individuals in a more fine-grained fashion.
Finally, when the $\varepsilon_i$ parameter (for the $i$-th generation) tends to zero, we avoid evaluating only those individuals that are behaviorally \emph{identical}.
However, since in the final generation the evolutionary process is expected to converge, the number of evaluations decreases again.


\section{Experimental setup}
\label{sec:experimental_setup}

We test our approach in three environments from the Atari Learning Environment implemented in OpenAI Gym \cite{bellemare13arcade, machado18arcade, brockman16openai}.
In particular, we use the Pong-v4, Bowling-v4, and Boxing-v4 environments, hereinafter simply referred to as Pong, Bowling and Boxing, respectively.
Table \ref{tab:params} shows the parameters used for the evolution. The parameters have been determined empirically, based on the knowledge gained in preliminary experiments. The parameter setting is the same for the three environments. The only difference is in the number of available actions for the decision module, that is $4$, $4$ and $6$ respectively for Pong, Bowling and Boxing.
It is important to note that the number of episodes is quite low.
This is due to the fact that in the fitness evaluation phase there is no learning involved, and thus the episodes are only used to evaluate the pipelines on average.
While a higher number of episodes would certainly increase the precision of the estimate of the performance of the pipeline in unseen episodes, it would significantly increase the computational cost of the search process.
All the runs were performed using an HPC that allocated $70$ CPUs and 4GB of RAM for each run, with a time limit of $6$ hours.
For each environment, we consider both the setting with and without frame-skipping, the first one being harder than the second one. For each environment and setting, we perform \NUMRUNS~runs. This number of runs is enough to ensure statistical significance since, given the results shown in Table \ref{tab:results}, the confidence interval ($95\%$) is low enough to validate our conclusions (see the next section for the results).
The confidence interval has been computed as $CI_{95\%} = \frac{t_{\{\NUMRUNS-1, 0.025\}}\sigma}{\sqrt{\NUMRUNS}}$, where $t$ is the critical value from the Student's t distribution, and $\sigma$ is the standard deviation.

\subsection{Image pre-processing}
Before feeding an image to the vision module, we perform the following pre-processing steps:
\begin{enumerate}[leftmargin=*]
    \item We remove the topmost 35 pixels: these pixels correspond to the part of the image that describes the ``status'' of the game, thus we remove it to avoid that the evolved pipelines use this information to take decisions.
    \item We resize the image to 96$\times$96: this operation speeds up the vision modules' computations, without losing information about the entities present in the game.
    \item We normalize the input in $[0,1]$ by performing a min-max normalization. 
\end{enumerate}

\subsection{Environments}

All the environments share the same type of observations, which consist in $210\times160$ RGB images, where each pixel is encoded with three 8-bit integers (one for each channel).

\subsubsection{Pong}
\label{subsubsec:pong}
The Pong environment (and its counterpart without frame-skipping, PongNoFrameskip) is a game in which there are two agents (here intended as ``players'', not to be confused with the agents that compose our pipelines), who play the Pong game.
Each agent controls a racket (which can move on the $y$ axis), and the goal of each agent is to send the ball farther than the opponent's $x$ position.\\
\textbf{Reward}: Each point scored by the playing agent (i.e., the green racket) gives a reward of 1 point.
On the other hand, each point scored by the opponent agent gives a reward of -1 points.
In all the other cases, the reward given to the agent is 0.\\
\textbf{Actions}: The action space consists of $4$ actions: two of them are NOP actions (i.e., they do not move the racket), one moves the racket upwards, and another moves the racket downwards.\\
\textbf{Termination criterion}: The simulation ends when either the agent or the opponent score $21$ points or after $10^5$ elapsed timesteps.

\subsubsection{Bowling}
\label{subsubsec:bowling}
The Bowling environment (and its counterpart BowlingNoFrameskip) consists in a bowling game, where the agent has to throw a ball to hit some pins.
Thus, differently from Pong this is a single-player game.\\
\textbf{Reward}: After each round (i.e., throwing twice the balls, or one in case of a strike), the agent receives a reward equal to the number of pins that have been hit.
Moreover, strikes and spares provide extra points (also given at the end of the round).
In all the other cases, the reward given to the agent is 0.\\
\textbf{Actions}: The agent can perform $4$ actions: a NOP action, an action that moves the ball upwards, one that moves the ball downwards, and an action that allows the agent to throw the ball.\\
\textbf{Termination criterion}: The simulation ends after $10$ rounds or after $10^5$ timesteps.
However, since an agent that does not know how to throw the ball will make the simulation become extremely time consuming, we reduced the amount of maximum timesteps to $5\times10^3$ during the evolutionary process.
Then, to obtain the results shown in Section \ref{sec:results}, we test the best pipelines evolved on the full task.

\subsubsection{Boxing}
\label{subsubsec:boxing}
In the Boxing (and its counterpart BoxingNoFrameskip) environment, there are two agents that compete: the white boxer (played by the agent) and the black boxer (which is the opponent).\\
\textbf{Reward}: When the agent hits the opponent, it can receive either 1 or 2 points, depending on the distance between the agents (hitting the opponent from a closer position gives more points).
On the other hand, when the agent is hit by the opponent it receives a reward that is the opposite of the one previously described.
In all the other timesteps, the reward given to the agent is 0.\\
\textbf{Actions}: The environment provides $18$ actions, composed of: NOP, movements in the $4$ cardinal directions, punching, and combinations of movements and punching (including diagonals).
However, here for simplicity we reduce the set of actions to the non-composite actions, i.e.: NOP, movements in the $4$ directions and punching.\\
\textbf{Termination criterion}: The simulation ends when either the agent or the opponent score $100$ points or after $10^5$ elapsed timesteps.

\begin{table}[ht!]
    \centering
    \caption{Parameters used in the experimentation.}
    \label{tab:params}
    \begin{tabular}{lc}
    \mytabrule
    \textbf{Parameter} & {\textbf{Value}} \\
    \mytabrule
    CMA-ES Population size ($p_c$) & 50 \\
    CMA-ES Initial mean & 0 \\
    CMA-ES Initial $\sigma$ & 0.1 \\
    \mytabrule
    GP Population size ($p_g$) & 50 \\
    GP Crossover probability & 0 \\
    GP Mutation probability & 1 \\
    GP Tournament size & 10 \\
    GP Elitism & Yes (1 elite) \\
    \mytabrule
    DBSCAN Number of samples (vision) ($n_{vm}$) & 100 \\
    DBSCAN Number of samples (decision) ($n_{dm}$) & 100 \\
    \mytabrule
    Number of generations ($g$) & 100 \\
    \mytabrule
    Maximum depth of decision tree & 4 \\
    Maximum depth of condition & 2 \\
    \mytabrule
    Number of convolutional kernels ($k$) & 2 \\
    Size of convolutional kernels & 5$\times$5$\times$3 \\
    \mytabrule
    Size of image & 96$\times$96 \\
    Number of episodes ($e$) & 3 \\
    \mytabrule
    Time limit & 6 hours \\
    Number of runs & \NUMRUNS \\
    \mytabrule
    \end{tabular}
\end{table}


\section{Experimental results}
\label{sec:results}

\begin{figure}[t]
    \centering
    \includegraphics[width=\columnwidth]{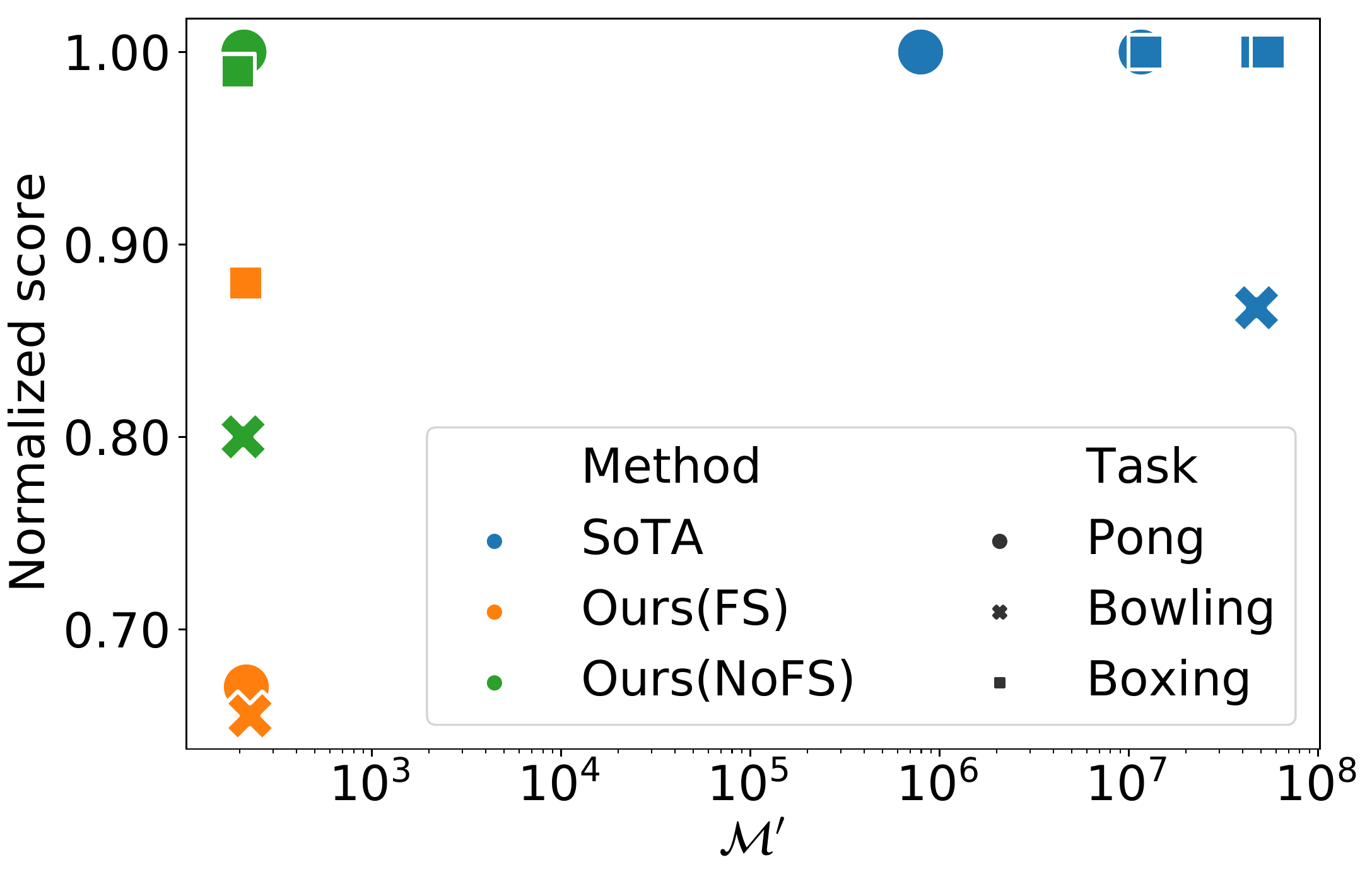}
    \caption{Comparison in terms of average normalized score across 100 episodes and $\mathcal{M'}$ (see Eq. \ref{eq:mprime}) between the best evolved pipelines and the state-of-the-art (SoTA) methods from the references reported in Table \ref{tab:results}.}
    \label{fig:pareto}
\end{figure}

The results obtained from the \NUMRUNS~available runs for each environment and setting are shown in Table \ref{tab:results}.
The results shown in the table have been obtained by testing the best evolved pipelines on 100 unseen episodes.
We observe that our approach performs well in the environments without frame-skipping, but it performs poorly in settings with frame-skipping.
While state-of-the-art approaches are able to achieve very good performance even in cases with frame-skipping, it is important to point out that these approaches are not interpretable, thus they do not provide any information about their inner processes.
On the other hand, while our approaches do not perform well when trained in setups with frame-skipping, they are completely transparent, potentially allowing an adaptation to domains with frame-skipping.
In fact, the non-interpretable approaches have $\mathcal{M'}$ scores (see Eq. \ref{eq:mprime}) in the order of $10^7$, while ours are in the order of $10^2$. This difference can be further appreciated in Figure \ref{fig:pareto}.
These results encourage future research in IRL, as adding more complexity to the pipelines would still yield a significant gain in interpretability w.r.t. the current non-interpretable state-of-the-art.

\begin{figure}[t]
    \centering
    \includegraphics[width=\columnwidth]{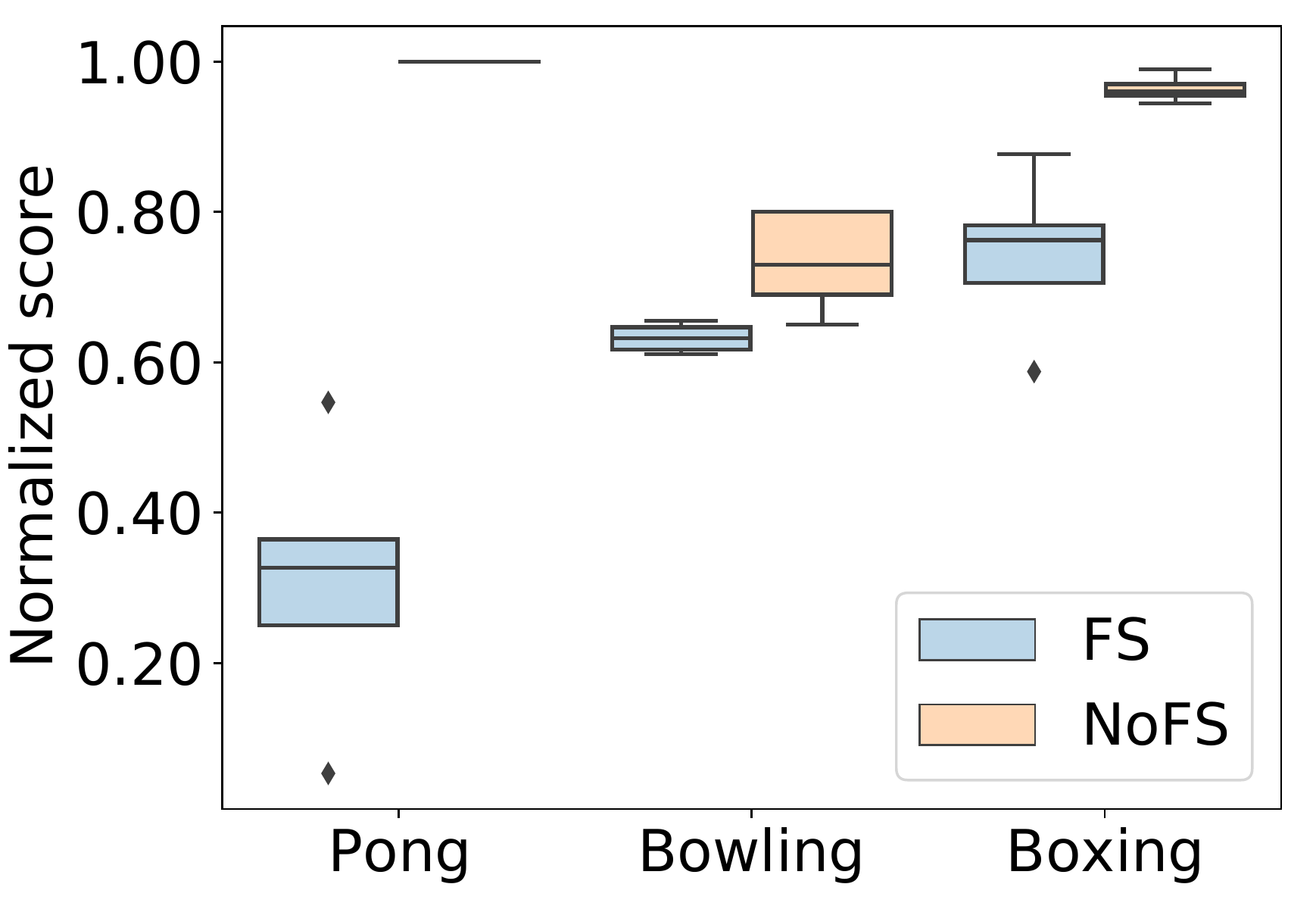}
    \caption{Distribution of the normalized scores of the best pipelines evolved in \NUMRUNS~runs of each setting for each environment, normalized w.r.t. the minimum and maximum scores allowed by each environment.}
    \label{fig:boxplot}
\end{figure}

Figure \ref{fig:boxplot} shows a comparison of the distribution of the scores of the best pipelines evolved in \NUMRUNS~runs of each environment and setting (normalized w.r.t. the minimum and maximum possible scores of the environments).
Once again we observe that, in all the cases, the settings without frame-skipping achieve very good performance (i.e., they are closer to one), while in the cases where frame-skipping is applied our algorithm is not able to achieve good performance.

\begin{table}[ht!]
    \centering
    \caption{Summary of the results of the best pipelines evolved in \NUMRUNS~runs of each setting for each environment. Each pipeline has been tested on 100 unseen episodes. ``FS'' stands for the setting with frame-skipping, ``NoFS'' indicates the setting without frame-skipping, ``SoTA'' indicates the results from the (non-interpretable) state-of-the-art. Please note that all the papers from the literature report results only in the FS setting.}
    \label{tab:results}
    \begin{tabularx}{\columnwidth}{llcccX}
        \mytabrule
        \textbf{Env.} & \textbf{Setup} & \textbf{Mean} & \textbf{Std.} & \textbf{Best} & \textbf{Reference}\\ \mytabrule
        \multirow{3}{*}{Pong} & FS (ours) & -6.97 & 8.49 & 7.42 & \\ 
        & NoFS (ours) & 21.00 & 0.00 & 21.00 \\
        & FS (SoTA) & - & - & 21.00 & \cite{wang2016dueling,salimans2017evolution, schrittwieser2020} \\
        \midrule
        \multirow{3}{*}{Bowling} & FS (ours) & 189.68 & 5.06 & 196.53 & \\
        & NoFS (ours) & 220.20 & 18.00 & 240.00 & \\
        & FS (SoTA) & - & - & 260.00 & \cite{schrittwieser2020,ecoffet2021} \\
        \midrule
        \multirow{4}{*}{Boxing} & FS (ours) & 48.59 & 19.05 & 75.37 & \\
        & NoFS (ours) & 92.78 & 3.09 & 98.00 & \\
        & FS (SoTA) & - & - & 100.00 & {\cite{bellemare13arcade, schrittwieser2020,horgan2018distributed,fortunato2017noisy,badia2020agent57,fan2022gdi,schrittwieser2021online}}\\
        \mytabrule
    \end{tabularx}
\end{table}

\begin{figure*}[htpb!]
    \hfill
    \hspace{-0.15cm}
    \begin{subfigure}{0.291\textwidth}
    \includegraphics[width=\columnwidth]{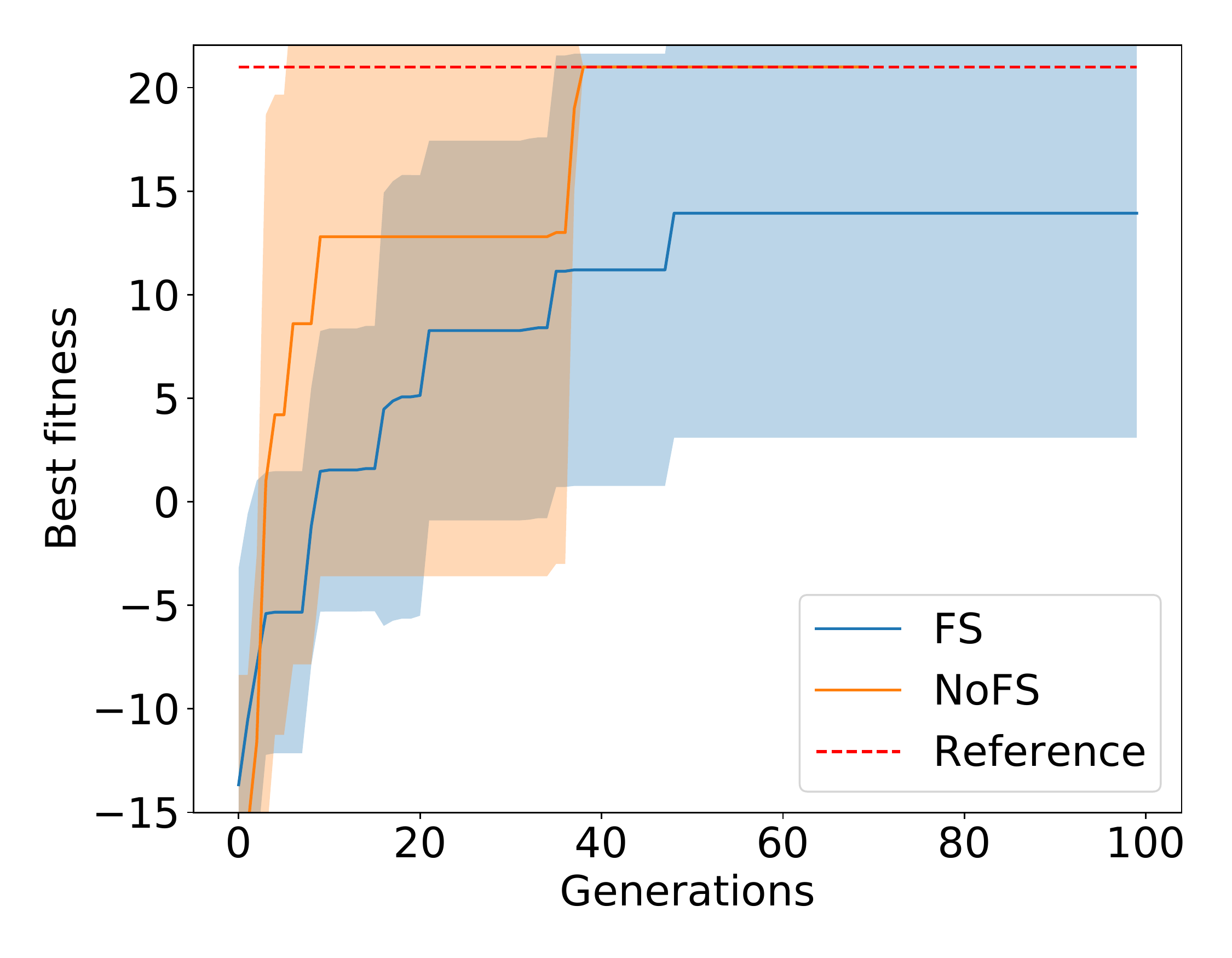}
    \caption{Fitness - Pong}
    \end{subfigure}
    \hfill
    \hspace{-0.15cm}
    \begin{subfigure}{0.291\textwidth}
    \includegraphics[width=\columnwidth]{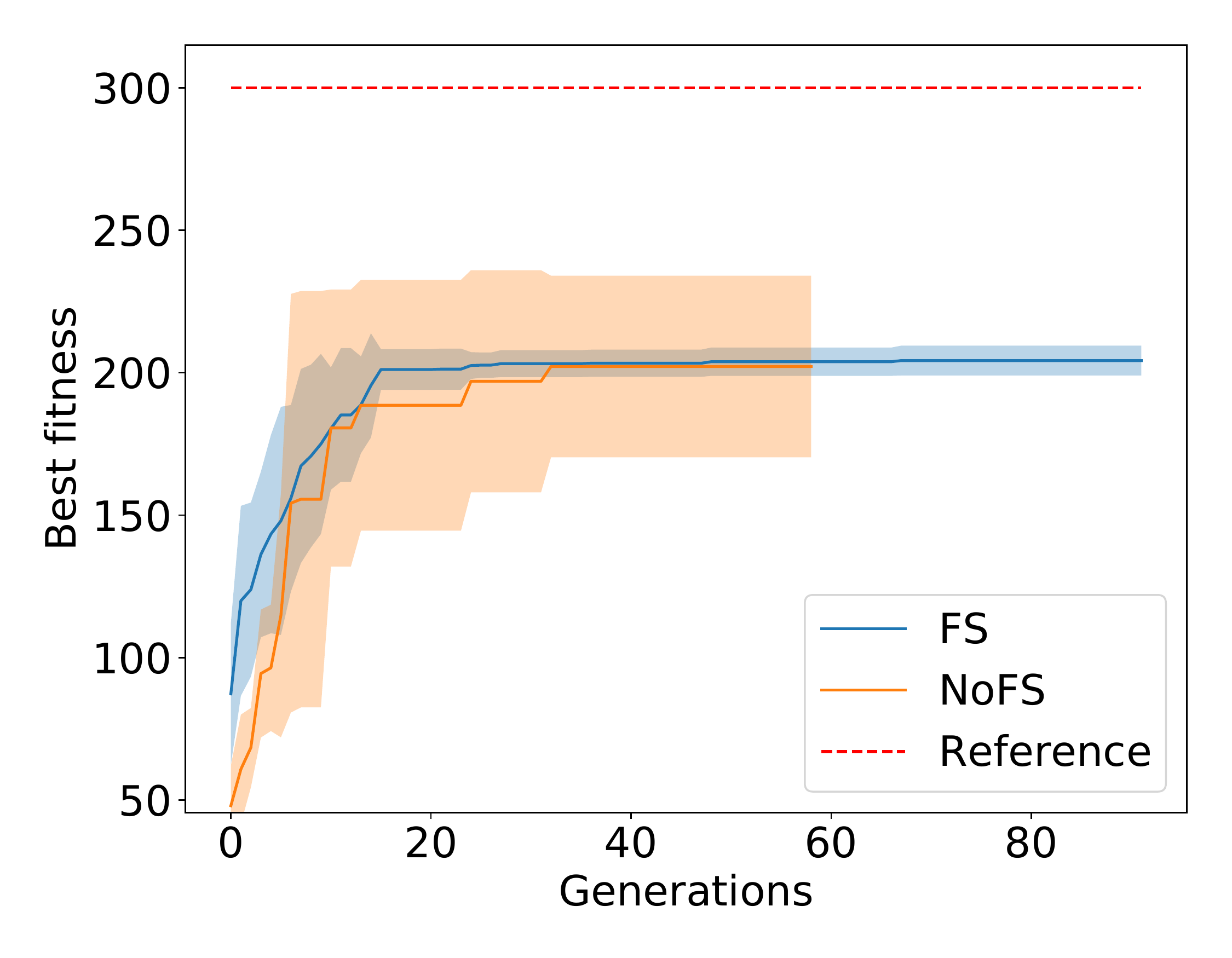}
    \caption{Fitness - Bowling}
    \end{subfigure}
    \hfill
    \begin{subfigure}{0.291\textwidth}
    \includegraphics[width=\columnwidth]{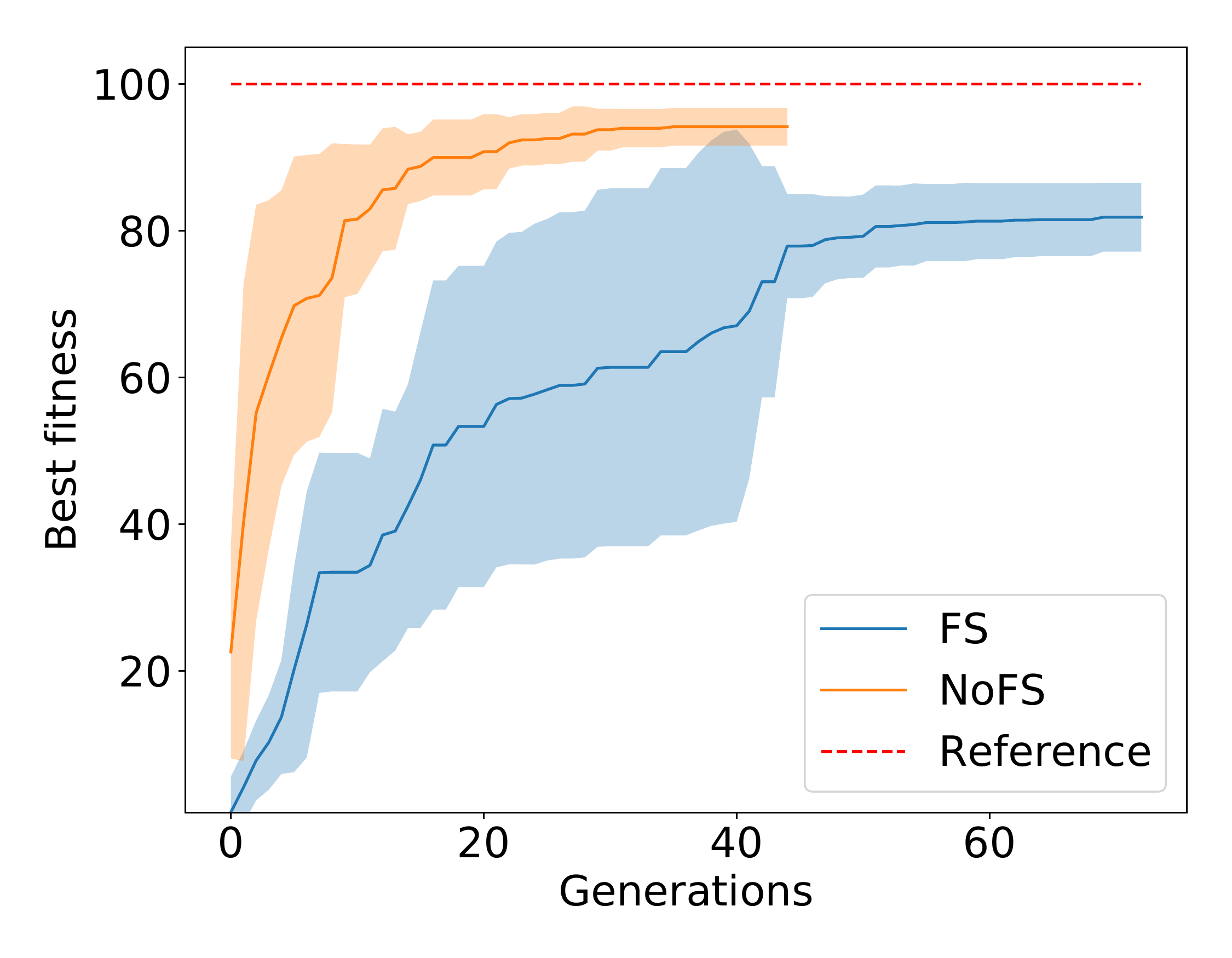}
    \caption{Fitness - Boxing}
    \end{subfigure}
    \hspace{0.06cm}
    \hfill
    \\
    \hfill
    \begin{subfigure}{0.318\textwidth}
    \includegraphics[width=\columnwidth]{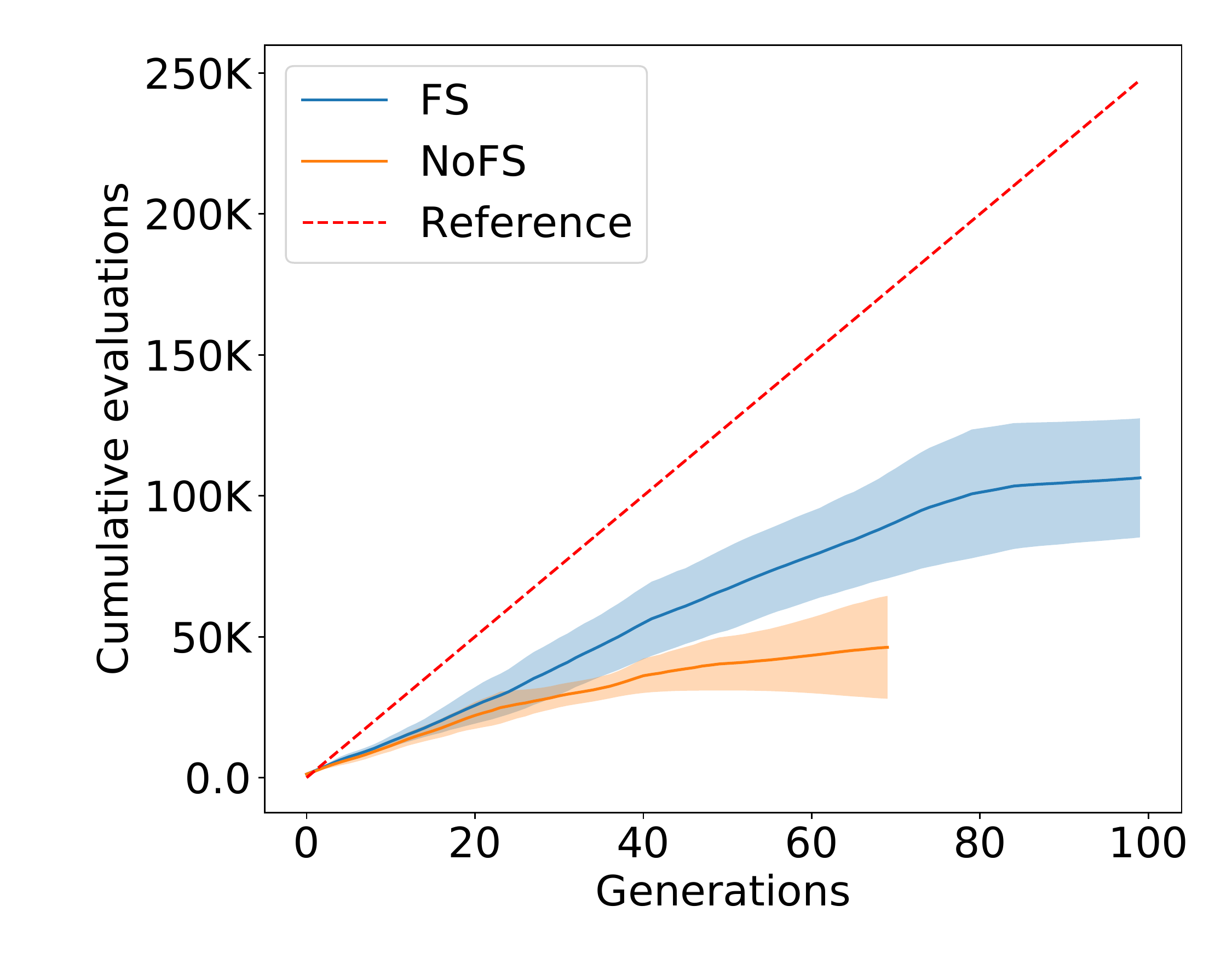}
    \caption{Evaluations - Pong}
    \end{subfigure}
    \hfill
    \begin{subfigure}{0.318\textwidth}
    \includegraphics[width=\columnwidth]{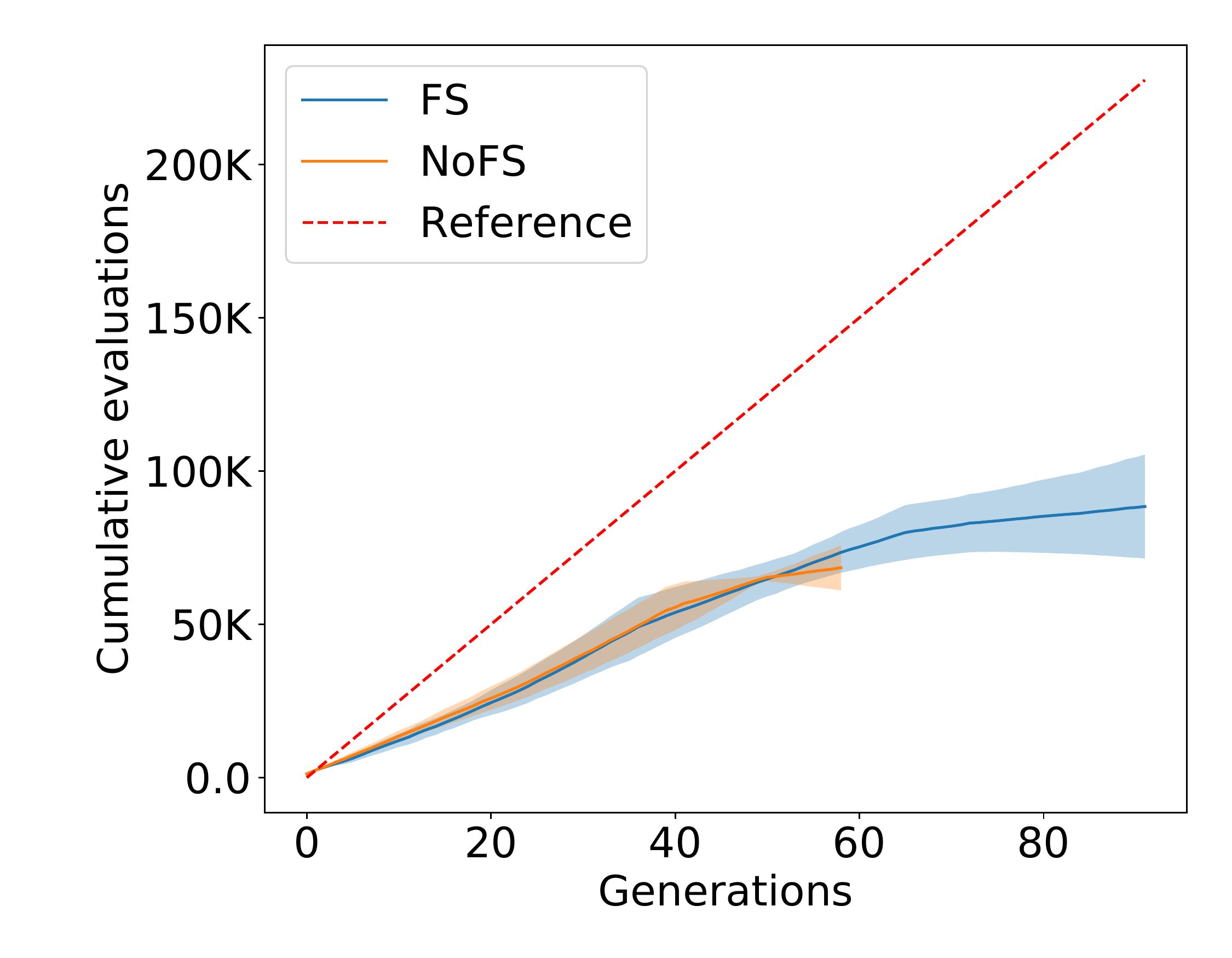}
    \caption{Evaluations - Bowling}
    \end{subfigure}
    \hfill
    \begin{subfigure}{0.318\textwidth}
    \includegraphics[width=\columnwidth]{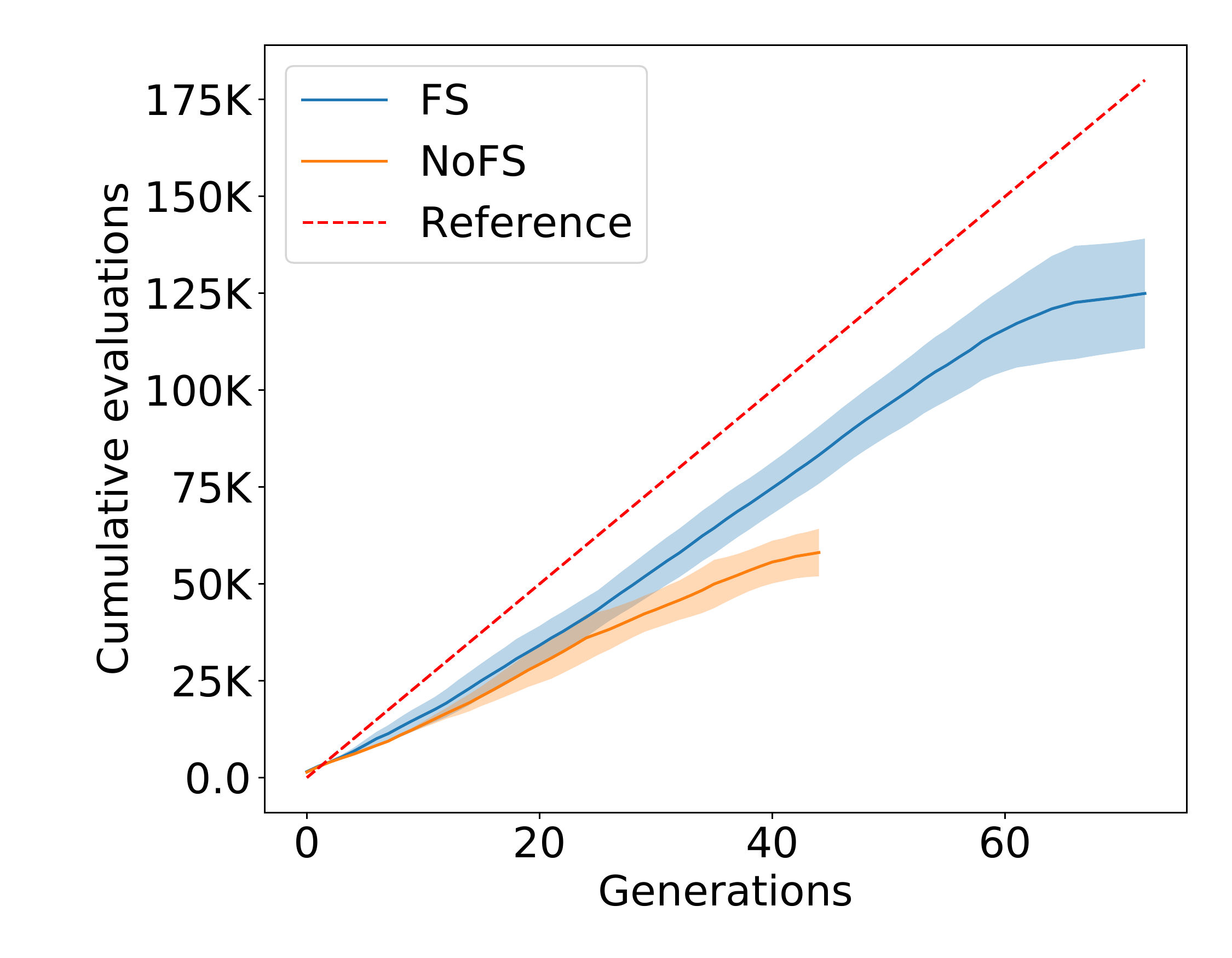}
    \caption{Evaluations - Boxing}
    \end{subfigure}
    \hfill
    \caption{Fitness and number of evaluations (mean $\pm$ std. dev. across \NUMRUNS~runs) over time. For the fitness, the reference is the best possible fitness allowed by the environment. For the number of evaluations, the reference is the number of evaluations needed if the clustering mechanism described in Section~\ref{subsubsec:reducing_the_number_of_evaluations} is not applied.}
    \label{fig:trends}
\end{figure*}

The fitness trends, as well the cumulative number of evaluations across generations (mean, represented as solid line, $\pm$ std. dev., represented as shaded area, across \NUMRUNS~runs) are shown for each environment and setting in Figure \ref{fig:trends}.
We observe that the clustering mechanisms allows us to save a significant amount of evaluations, increasing the efficiency of the co-evolutionary process (approximately saving 41\%$\pm$12\% evaluations).

\subsection{Analysis of the best evolved pipelines}\label{subsec:analysis}

We conclude our presentation of the results with an analysis of the best evolved pipelines obtained by means of the proposed method.
Note that, since the settings without frame-skipping produced better results, we will limit our analysis to these settings.

The best pipelines obtained are shown graphically in Figure \ref{fig:bests}.
In order to improve the readability, the decision modules have been manually simplified, deleting the conditions that always evaluate to the same truth value.

\subsubsection{Pong}
\label{subsubsec:pong_results}
\begin{figure*}[t]
    \hfill
    \begin{subfigure}[b]{0.30\textwidth}
        \includegraphics[width=0.9\columnwidth]{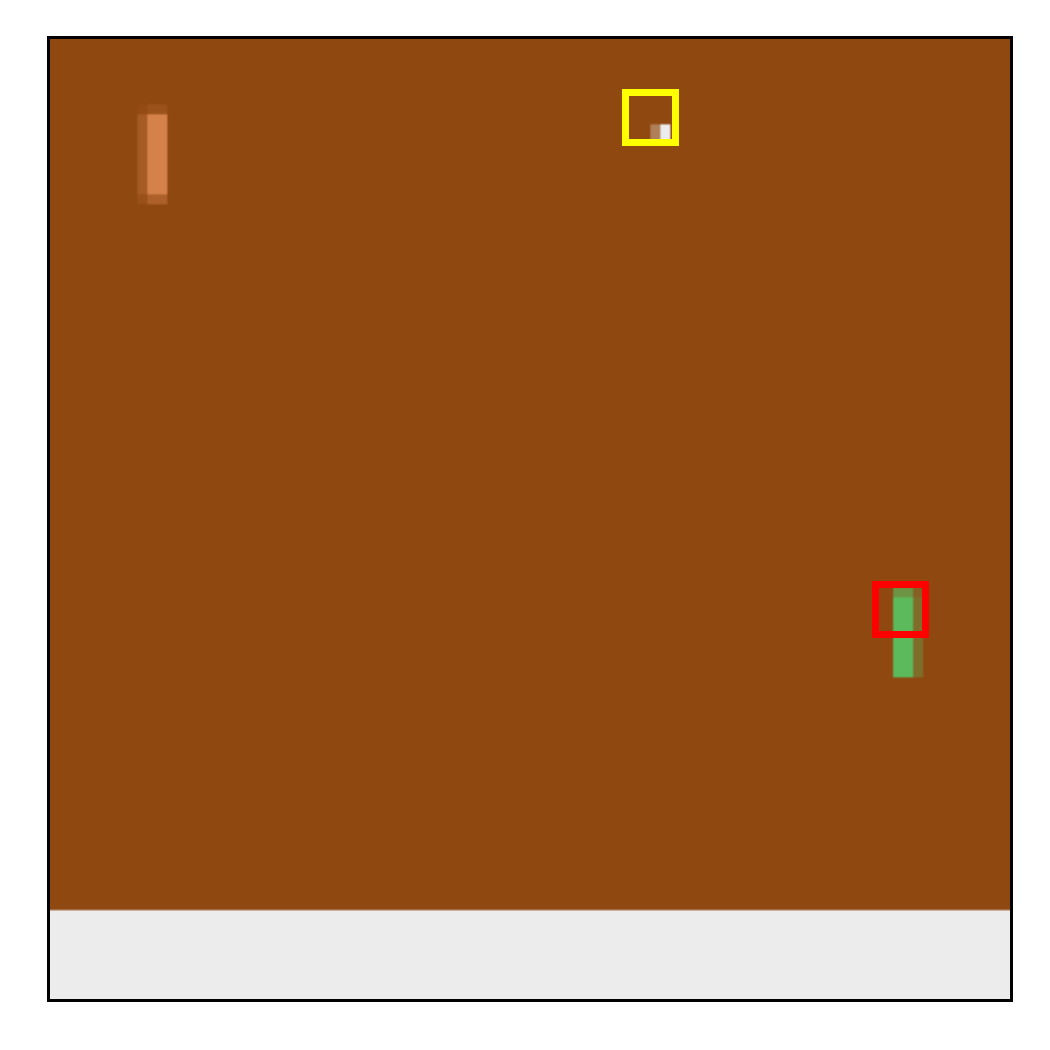}
    \caption{Vision module - Pong}
    \label{fig:vm_pong}
    \end{subfigure}
    \hfill
    \begin{subfigure}[b]{0.30\textwidth}
        \includegraphics[width=0.9\columnwidth]{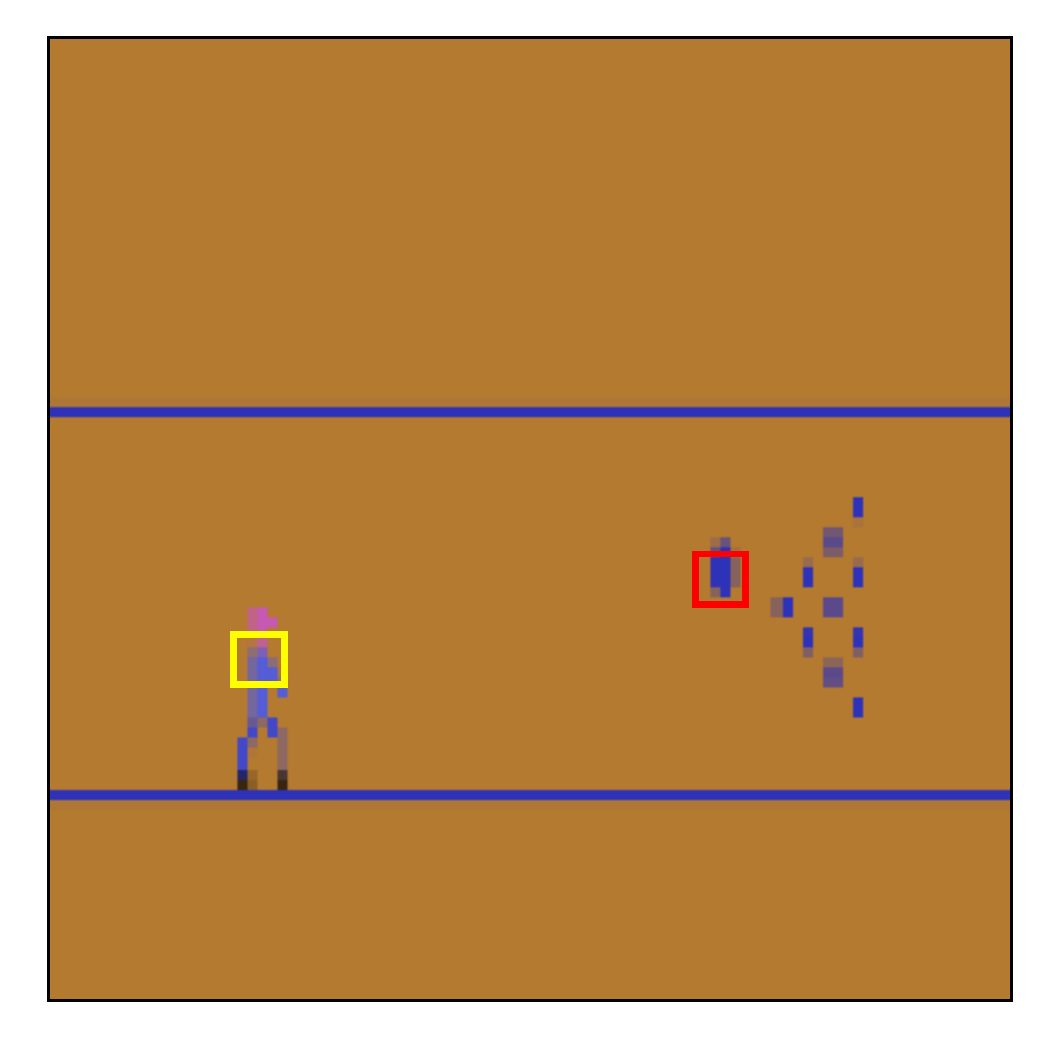}
    \caption{Vision module - Bowling}
    \label{fig:vm_bowling}
    \end{subfigure}
    \hfill
    \begin{subfigure}[b]{0.30\textwidth}
        \includegraphics[width=0.9\columnwidth]{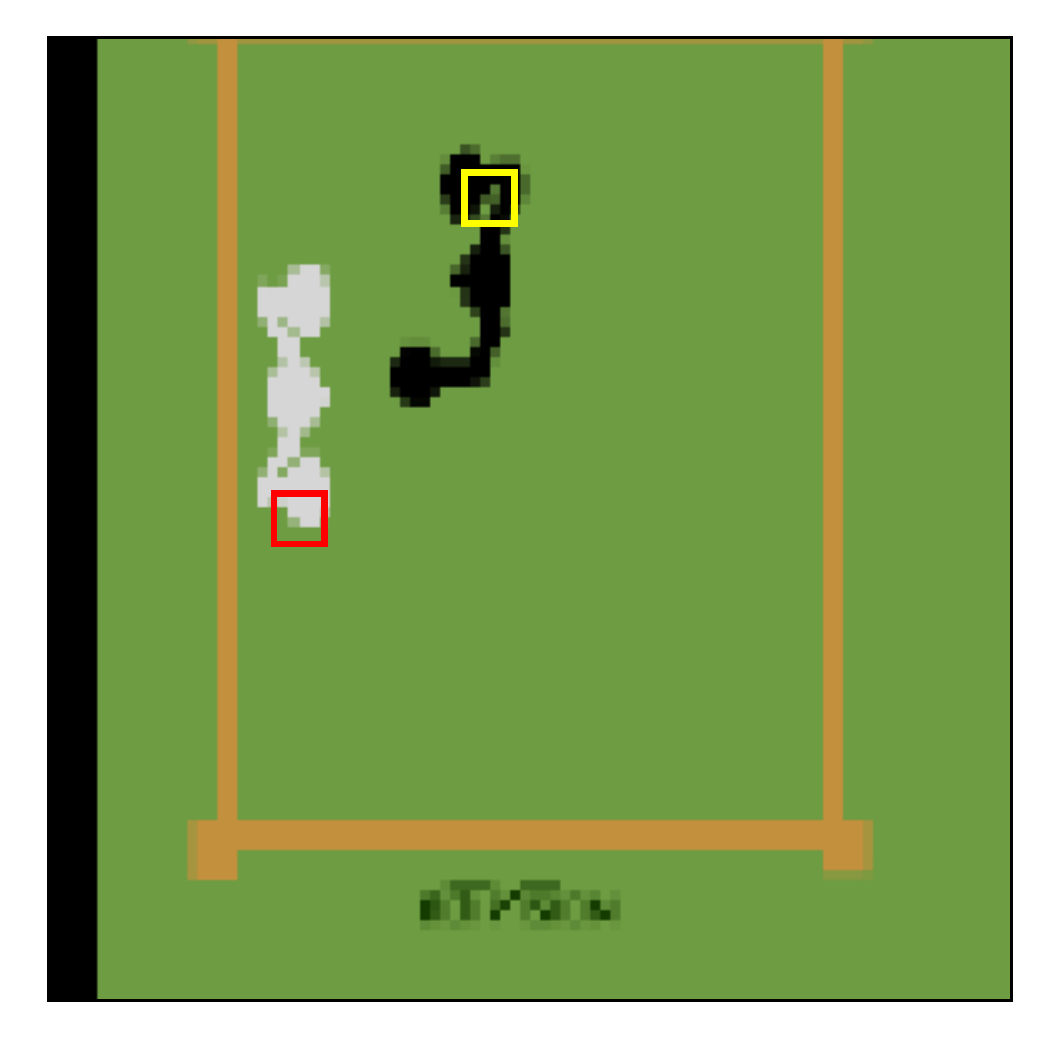}
    \caption{Vision module - Boxing}
    \label{fig:vm_boxing}
    \end{subfigure}
    \hfill
    \\
    \hfill
    \begin{subfigure}[b]{0.30\textwidth}
    \resizebox{0.9\columnwidth}{!}{
    \begin{tikzpicture}[transform shape]
        \node [box] (root) {$53.9 > y_r$};
        \node [box, below=-0cm of root, xshift=-1.35cm] (l) {$y_b < x_r$};
        \draw (root) -| (l) node [midway, above] () {T};
        \node [box, below=-0.1cm of root, xshift=+1.35cm] (r) {$87.4 > y_b$};
        \draw (root) -| (r) node [midway, above] () {F};
        \node [leaf, below=0.16cm of l, xshift=-0.85cm] (ll) {$DOWN$};
        \draw (l) -| (ll) node [midway, above] () {T};
        \node [leaf, below=0.16cm of l, xshift=+0.85cm] (lr) {$NOP$};
        \draw (l) -| (lr) node [midway, above] () {F};
        \node [leaf, below=-0cm of r, xshift=-1cm] (rl) {$UP$};
        \draw (r) -| (rl) node [midway, above] () {T};
        \node [leaf, below=-0cm of r, xshift=+1cm] (rr) {$NOP$};
        \draw (r) -| (rr) node [midway, above] () {F};
    \end{tikzpicture}
    }
    \caption{Decision module - Pong}
    \label{fig:dm_pong}
    \end{subfigure}
    \hfill
    \begin{subfigure}[b]{0.30\textwidth}
    \resizebox{0.9\columnwidth}{!}{
     \begin{tikzpicture}[transform shape]
        \node [box] (root) {$y_p > 48.5$};
        \node [leaf, below=-0cm of root, xshift=-1.5cm] (l) {$UP$};
        \draw (root) -| (l) node [midway, above] () {T};
        \node [box, below=-0.1cm of root, xshift=+1.5cm] (r) {$x_p > x_b$};
        \draw (root) -| (r) node [midway, above] () {F};
        \node [leaf, below=-0.1cm of r, xshift=-1.5cm] (rl) {$DOWN$};
        \draw (r) -| (rl) node [midway, above] () {T};
        \node [leaf, below=-0.1cm of r, xshift=+1.5cm] (rr) {$THROW$};
        \draw (r) -| (rr) node [midway, above] () {F};
    \end{tikzpicture}
    }
    \caption{Decision module - Bowling}
    \label{fig:dm_bowling}
    \end{subfigure}
    \hfill
    \begin{subfigure}[b]{0.30\textwidth}
    \resizebox{0.9\columnwidth}{!}{
    \begin{tikzpicture}[transform shape]
        \node [box] (root) {$y_p < y_o$};
        \node [leaf, below=-0cm of root, xshift=-1.5cm] (l) {$PUNCH$};
        \draw (root) -| (l) node [midway, above] () {T};
        \node [leaf, below=-0.1cm of root, xshift=+1.5cm] (r) {$UP$};
        \draw (root) -| (r) node [midway, above] () {F};
    \end{tikzpicture}
    }
    \caption{Decision module - Boxing}
    \label{fig:dm_boxing}
    \end{subfigure}
    \hfill
    \caption{Best evolved pipelines for the three environments (without frame-skipping). On the top, we show the entities discovered by the vision modules, while on the bottom we show the corresponding decision modules.}
    \label{fig:bests}
\end{figure*}
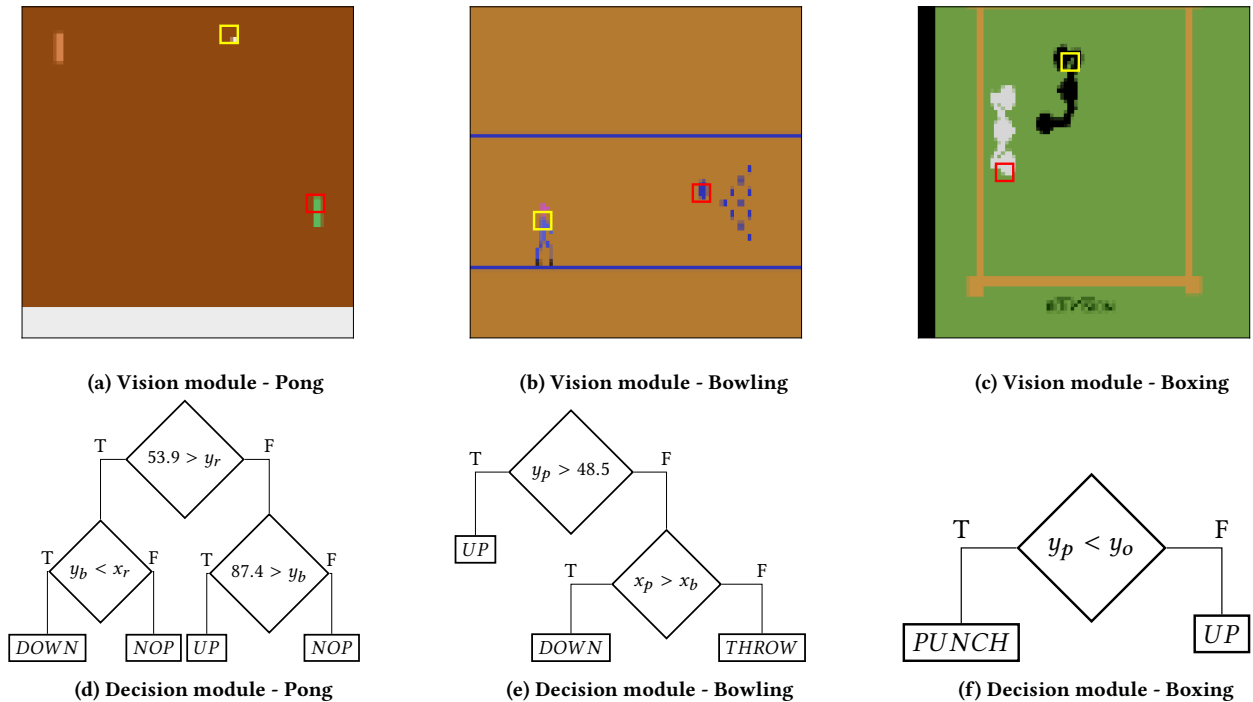

As shown in Figure \ref{fig:vm_pong}, the vision module of the best pipeline evolved for Pong detects the two most important entities: The racket and the ball, giving in output their coordinates: $x_r, y_r, x_b, y_b$, where the subscript $r$ refers to the player's racket and the subscript $b$ refers to the ball.

The policy of the decision-making module for this environment, as shown in Figure \ref{fig:dm_pong}, works as follows.
First of all, it checks whether the racket is on the upper part of the screen.
If so, it checks whether the y-coordinate of the ball is less than the x-coordinate of the racket.
This condition can be simplified: the horizontal position of the racket is constant ($x_r = 82$).
So, this condition is equivalent to $y_b < 82$, which means that the ball is not near to the bottom wall (white part in Figure \ref{fig:vm_pong}).
Then, if this condition evaluates to True, the decision module decides to go downwards, otherwise if does not perform any action.
On the other hand, when the racket is on the lower part of the screen, the decision module checks $87.4 > y_b$, i.e., if the ball is not near to the bottom wall.
If so, it decides to go upwards, otherwise it does not perform any action.

\subsubsection{Bowling}
\label{subsubsec:bowling_results}
The entities recognized by the vision module are shown in Figure \ref{fig:vm_bowling}, and they are: the person throwing the ball and the ball.
Thus, the coordinates that the vision module sends to the decision module are: $x_p, y_p, x_b, y_b$, where the subscript $p$ refers to the person, and the subscript $b$ refers to the ball.

The decision module (Figure \ref{fig:dm_bowling}) performs the following decision-making process.
First of all, it checks the vertical coordinate of the person and, if it is too low on the screen ($yp > 48.5$, note that the top-left screen has coordinates $(0, 0)$), it moves the person upwards, so that it is positioned correctly to perform a good shot.
Then, after the person is correctly positioned on the bowling alley, it checks another condition to understand whether the ball has been thrown or not ($x_p > x_b$).
In fact, if the ball has not been thrown, it stands behind the person.
Thus, if the ball is moving toward the pins, it moves the position of the ball downwards.
Otherwise, if the ball is still in the hand of the person, it makes the person throw the ball.

\subsubsection{Boxing}
\label{subsubsec:boxing_results}
In Figure \ref{fig:vm_boxing}, we show the two entities recognized by the vision module: the punch of the player, and the position of the right arm of the opponent.
Thus, the vision module returns their coordinates: $x_p, y_p, x_o, y_o$, where the subscript $p$ refers to the player and the subscript $o$ refers to the opponent.
Moreover, by testing the vision module, we have observed that the kernel does not work perfectly, i.e., it happens that the kernel that should recognize the position of the player recognizes a part of the opponent.
Nevertheless, this pipeline manages to achieve a near optimal average score (on unseen episodes) of $98$, i.e., $1\%$ away from the maximum.

The decision module, shown in \ref{fig:dm_boxing}, works as follows.
If the player is positioned upper than the opponent (again, note that the top-left corner has coordinates $(0, 0)$), it tries to punch the opponent, otherwise it goes upwards, to reach the opponent.

Interestingly, such a simple policy (encoded by the decision module) allows us to understand some properties about the pipeline and the game itself.
First, the game can be played with very good results by using only two actions (note that the environment provides 18 actions for the player).
Moreover, the other actions are not needed because the opponent chases the player.
For this reason, the decision module finds more advantageous to use a semi-defensive strategy, i.e., always punch if the opponent is reachable by the punches, and chase it only when it is upwards.


\section{Conclusions and future works}
\label{sec:conclusions}

Reinforcement learning (RL) has made significant progresses in recent years.
However, mainstream RL methodologies, typically based on deep learning, are very hard to understand.
In this paper, we proposed a novel methodology (based on a kind of divide-et-impera paradigm) for evolving interpretable systems for RL tasks with visual inputs. In particular, our approach is based on pipelines characterized by a separation of concerns between a vision module (which uses convolutional kernels) and a decision module (based on a decision tree).
Our results show that our approach is able to learn how to effectively play three Atari games in simplified settings (i.e., without frame-skipping).
However, when applying frame-skipping to the environments, our approach is not able to achieve satisfactory performance.

Future work should introduce ways to address the uncertainty in non-deterministic settings (i.e., with frame-skipping), in order to make this approach more robust to noise in the environment and achieve performances comparable to those of the state-of-the-art algorithms developed for these settings. In this sense, two possibilities would be to incorporate in our approach some mechanisms used in evolutionary optimization in the presence of noise \cite{arnold2002noisy}, or using fuzzy \cite{janikow1998fuzzy} or probabilistic \cite{lakshminarayanan2016decision} decision trees.




\clearpage

\newcommand{\showDOI}[1]{\unskip}

\bibliographystyle{ACM-Reference-Format}
\bibliography{main,sample-base}



\end{document}